%% file: main.tex
\documentclass[11pt]{article}

\usepackage[preprint]{acl}

\usepackage{times}
\usepackage{latexsym}
\usepackage{array}
\usepackage{amsmath}
\usepackage{multirow}

\usepackage{amssymb}
\usepackage[T1]{fontenc}
\usepackage[utf8]{inputenc}

\usepackage{microtype}
\usepackage{booktabs}
\usepackage{inconsolata}

\usepackage{graphicx}
\usepackage{colortbl}
\PassOptionsToPackage{table}{xcolor}
\PassOptionsToPackage{dvipsnames}{xcolor}

\definecolor{hj_blue}{HTML}{B3C8CF}
\definecolor{hj_gray}{HTML}{BFCFE7}
\definecolor{hj_deep}{HTML}{749BC2}

\usepackage[most]{tcolorbox}

\title{NL2Scratch: An Executable Benchmark and Evaluation for \\ Block-Based Programming}

\author{
    Heejin Do\thanks{Equal contribution.}$^{1, 2}$ \quad
    Alexandre Ballenghien\footnotemark[1]$^{1}$ \quad
    Yang Wu$^{1}$ \quad
    {April Yi Wang$^{1}$} \\ \text{} \\
  $^{1}$Department of Computer Science, ETH Zurich \quad \\
  $^2$ ETH AI Center \\
 \texttt{heejin.do@ai.ethz.ch, aballenghien@student.ethz.ch} \\
 \texttt{\{yang.wu, april.wang\}@inf.ethz.ch}}

\usepackage{xcolor}
\usepackage[table]{xcolor}

\definecolor{promptnavy}{HTML}{2B4F7E}   
\definecolor{promptblue}{HTML}{4A7BB7}   
\definecolor{promptsky}{HTML}{EEF4FA}  

\newtcolorbox{promptbox}[1]{
  enhanced,
  colback=promptsky,
  colframe=promptnavy,
  colbacktitle=promptnavy,
  coltitle=white,
  fonttitle=\bfseries,
  title=#1,
  arc=2mm,
  boxrule=0.8pt,
  left=3mm, right=3mm, top=2mm, bottom=2mm,
  breakable,
  before skip=8pt,
  after skip=8pt,
}

\newcommand{\role}[1]{\textbf{\color{promptnavy}[#1]}}

\begin{document}
\maketitle

\input{Contents/00_abstract}

\input{Contents/01_introduction}
\input{Contents/02_related_work_hj}
\input{Contents/03_method}
\input{Contents/04_experiments}

\input{Contents/05_results_hj}
\input{Contents/08_Conclusion}

\input{Contents/07_Limitations}
\input{Contents/08_ethical}

\bibliography{custom}

\appendix
\input{Contents/09_appendix}

\end{document}

%% file: Contents/00_abstract.tex
\begin{abstract}

Block-based programming environments such as Scratch are widely used in early programming education, yet natural-language-to-code (NL2Code) research has focused primarily on text-based languages. Scratch programs are event-driven, visually compositional, and distributed across concurrent scripts, making conventional NL2Code assumptions and evaluation insufficient. We introduce NL2Scratch\footnote{\url{https://github.com/doheejin/NL2Scratch}}, an executable benchmark for natural-language-to-Scratch generation comprising 311,648 parser-valid NL--program pairs, whose program side is extracted from real Scratch projects and paired with semantically aligned NL descriptions. For reliable evaluation beyond surface overlap, we propose Semantic Alignment Consistency (SAC), an interpretable slot-level metric for measuring semantic agreement between descriptions and programs. With SAC, we construct a  semantically validated pool of 23{,}594 examples, and a slot-balanced 800 diagnostic benchmark. Experiments across instruction-tuned and fine-tuned LLMs reveal a notable gap between lexical similarity and semantic alignment: models achieving token-level F1 above 0.93 often fail to attain perfect SAC, particularly on longer examples. Errors concentrate on operational slots like actions, conditions, and numeric arguments, exposing failure modes largely invisible under conventional metrics.
\end{abstract}

%% file: Contents/01_introduction.tex
\section{Introduction}

Block-based programming environments such as Scratch~\cite{resnick2009scratch,maloney2010scratch} play a central role in early programming education~\citep{brennan2012new}, especially in K-12 and informal learning contexts. Yet modern natural-language-to-code (NL2Code) research has focused almost exclusively on text-based programming languages such as Python, SQL, and Java~\citep{chen2021codex,austin2021program,
yu2018spider,hendrycks2021apps,lu2021codexglue,roziere2023codellama}. Scratch fundamentally differs from conventional programming languages: programs are event-driven, visually compositional, and distributed across multiple concurrent scripts. These differences challenge core assumptions underlying existing NL2Code methods and raise the question of whether current evaluation protocols transfer reliably to block-based programming.

Existing NL2Code systems and benchmarks largely assume that programs are sequential textual artifacts whose correctness can be approximated through token overlap~\citep{papineni2002bleu,ren2020codebleu} or unit-test execution~\citep{chen2021codex,austin2021program,hendrycks2021apps}. In Scratch, however, correctness depends on structural coordination across scripts: event handling, shared state, broadcast synchronization, and nested control structures. Thus, programs with high surface similarity may still fail to execute or behave incorrectly, making standard NL2Code evaluation unreliable in block-based settings.

Despite growing interest in AI for programming education~\citep{kazemitabaar2023studying,finnieansley2022robots}, recent Scratch-focused work has primarily explored how AI and off-the-shelf large language models can support learners around Scratch programming. Prior systems provide conversational tutoring and scaffolded guidance~\citep{chen2024chatscratch,druga2025scratch}, repair~\citep{fein2025litterbox,si2025viscratch}, program understanding~\citep{si2026scratcheval}, GUI interaction~\citep{zhang2026see} rather than executable program synthesis from natural language, and automatically generated comprehension questions for existing Scratch programs \cite{obermuller2025automatically}. 
These works demonstrate the promise of AI-assisted Scratch learning, however, to our knowledge, no benchmark provides executable and semantically validated NL–Scratch pairs for generation.

To address this gap, we introduce NL2Scratch, an executable benchmark for natural-language-to-Scratch generation. NL2Scratch contains 311,648 parser-valid NL--program pairs whose program side is extracted from real Scratch projects and whose natural-language side is generated through a semantics-preserving description and rewriting pipeline. We further construct a semantically validated pool of 23,594 examples and a slot-balanced 800-example diagnostic benchmark covering diverse Scratch-specific structures. To enable scalable semantic verification, we propose Semantic Alignment Consistency (SAC), an interpretable slot-level framework that measures semantic agreement between NL descriptions and Scratch programs across eleven behavioral slots, including event triggers (e.g., green-flag or key-press events), actions (e.g., motion or variable updates), and control structures (e.g., loops and conditionals). SAC serves both as a semantic verification signal for benchmark construction and as a fine-grained diagnostic for model evaluation.

Leveraging NL2Scratch and SAC, we conduct an empirical study across instruction-tuned and fine-tuned LLMs, including proprietary models (GPT-4.1 and GPT-4o-mini) and open-weight models (FLAN-T5, Qwen2.5-7B-Instruct, and Llama-3.1-8B-Instruct). Our analysis reveals a consistent gap between conventional generation metrics (e.g., F1, Parser Success) and semantic alignment: even models achieving token-level F1 and parse success rates above 0.93 attain perfect semantic alignment on fewer than 18.9\% of examples, with the gap widening as program complexity increases. Slot-level analysis further shows that errors concentrate on operational components, including actions, conditions, and numeric arguments, rather than event or identifier slots, exposing failure modes largely invisible under conventional NL2Code metrics. Finally, we demonstrate that SAC can be used beyond evaluation as an inference-time selection signal. Across open-weight models, SAC-based reranking substantially improves alignment with reference programs, increasing perfect SAC$_{\text{ref}}$ alignment by up to 32.6 percentage points without retraining.

\begin{enumerate}
    \item We introduce {NL2Scratch}, the first executable benchmark for natural-language-to-Scratch generation, constructed from real Scratch projects and validated through parser-level and semantic consistency checks.

    \item We propose {Semantic Alignment Consistency (SAC)}, an interpretable slot-level verification procedure that enables reliable evaluation beyond surface-form overlap.

    \item We show that robust LLMs, under both in-context learning and supervised fine-tuning, exhibit a notable gap between surface similarity and structural accuracy, revealing failure modes that remain largely invisible under conventional metrics.
\end{enumerate}

%% file: Contents/02_related_work_hj.tex
\section{Related Work}

\paragraph{Natural Language to Code Generation.}

NL2Code maps natural-language specifications to programs and has advanced rapidly with code-specialized LLMs such as Code Llama~\citep{roziere2023codellama}, StarCoder~\citep{li2023starcoder}, and DeepSeek-Coder~\citep{guo2024deepseek}. Existing benchmarks derive NL--program pairs from a variety of sources, including human-written programming problems and reference solutions~\citep{chen2021codex,austin2021program,lu2021codexglue,yu2018spider,hendrycks2021apps}. They primarily target textual languages such as Python and Java, and evaluate generations through lexical similarity (e.g., BLEU~\citep{papineni2002bleu}, CodeBLEU~\citep{ren2020codebleu}) or execution-based success with unit tests. Unlike conventional settings, Scratch programs are event-driven and structurally compositional, making correctness dependent on triggers, control flow, and argument values. Accordingly, our NL2Scratch combines benchmark construction from program artifacts with explicit semantic validation during both dataset curation and evaluation.

\paragraph{Modeling Block-Based Programming Languages.}

Block-based programming environments such as Scratch~\citep{resnick2009scratch}, Snap!~\citep{harvey2013snap}, and Blockly~\citep{blocky2015trower} are widely used in introductory programming education. Unlike textual languages, programs are expressed through visually composable blocks and event-driven interactions across multiple scripts, often involving broadcasts, shared variables, and multimodal assets such as costumes and backdrops~\citep{maloney2010scratch,fraser2021litterbox}. These properties make program behavior highly sensitive to structural elements such as triggers, control flow, and referenced entities. Prior work has explored token-based~\citep{griebl2023applicability}, AST-inspired~\citep{fein2022code2vec}, and graph-based representations~\citep{wu2021hybrid} for Scratch programs. However, little is known about how modern LLMs perform on executable natural-language-to-Scratch generation, a setting that NL2Scratch is designed to study.

\paragraph{AI Support for Learning Scratch.}

A growing body of work applies AI techniques to Scratch-based learning environments. Prior research has explored conversational tutoring systems that assist learners through dialogue and scaffolded guidance~\citep{chen2024chatscratch}, automated debugging and program repair tools~\citep{fein2025litterbox,si2025viscratch}, and program-understanding benchmarks that evaluate models' ability to interpret Scratch projects~\citep{si2026scratcheval}. More recent agent-based approaches investigate interaction within graphical programming environments~\citep{zhang2026see}. These efforts primarily treat Scratch programs as objects to understand, analyze, or modify. In contrast, our focus is executable natural-language-to-Scratch generation. To our knowledge, NL2Scratch is the first benchmark to provide large-scale NL--program pairs derived from real Scratch projects together with aligned quality verification.

%% file: Contents/03_method.tex
\begin{figure*}[t]
    \centering
    \includegraphics[width=\textwidth]{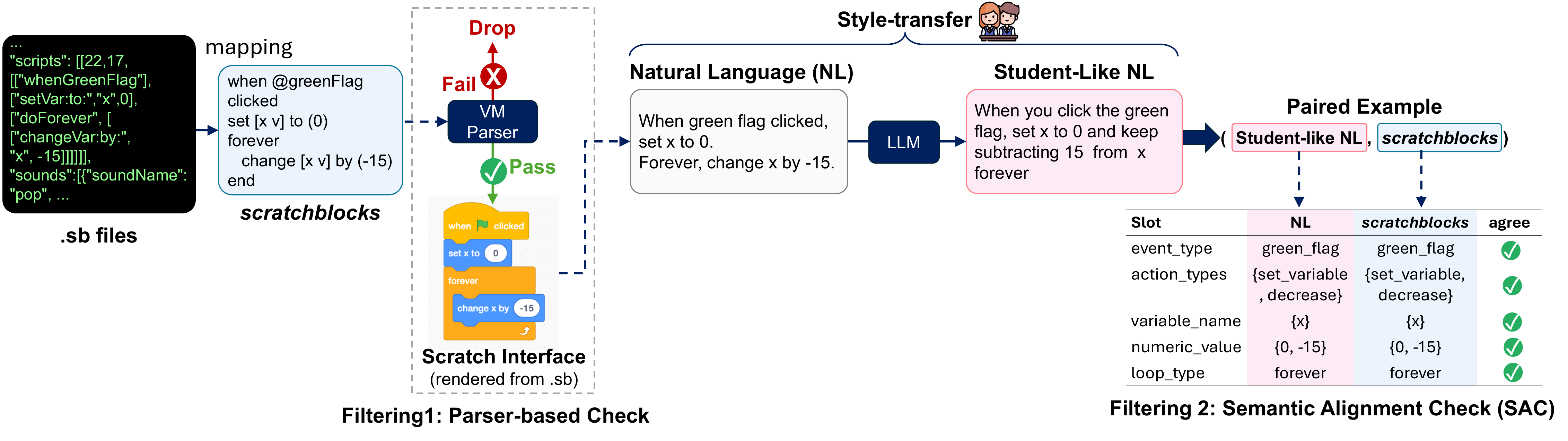}
    \caption{\textbf{NL2Scratch data construction pipeline.} Raw Scratch projects (.sb files) are converted into normalized \textit{scratchblocks}. The outputs are then validated through two complementary filters: (1) parser-based validation, which ensures structural validity and renderability as executable Scratch Interface, and (2) SAC, which verifies semantic agreement between NL descriptions and programs across behavioral slots. Only high-consistency pairs are retained, yielding aligned (student-like NL description, Scratch program) pairs.
    }
    \label{fig:framework}
\end{figure*}

\section{NL2Scratch}

NL2Scratch is a benchmark for natural-language-to-Scratch generation designed to capture the structural characteristics of block-based programming. Unlike conventional NL2Code benchmarks over sequential textual programs, Scratch programs are event-driven, visually compositional, and often distributed across multiple concurrent scripts. Consequently, benchmark construction must ensure both structural validity on the program side and semantic fidelity between natural-language descriptions and executable behavior.

To this end, we construct NL2Scratch through three stages: (i) \textit{scratchblocks} extraction and validation, (ii) natural-language generation, and (iii) semantic consistency verification. The resulting dataset contains 311,648 parser-valid NL--program pairs derived from real Scratch projects. We further use semantic verification to construct a high-confidence evaluation pool and an 800-example diagnostic evaluation set for controlled analysis.

\subsection{Task Definition}

Given a natural-language description $x$ written in a student-facing style, the goal is to generate a Scratch program $\hat{p}$ that implements the described behavior. We represent target Scratch programs using \textit{scratchblocks}, a textual notation for Scratch blocks that provides a compact and human-readable representation while preserving program structure. In this representation, blocks are linearized line-by-line and hierarchical control structures are expressed through indentation.

Formally, each example is a pair $(x,p)$ consisting of a natural-language description $x$ and a Scratch program representation $p$. Unlike conventional NL2Code tasks, Scratch programs encode event-triggered execution, nested control flow, repeated behaviors, state updates, and interactions between sprites. Consequently, lexical overlap alone is insufficient for evaluation: programs with high token similarity may still differ substantially in their intended semantics due to incorrect triggers, actions, conditions, or argument values.

\subsection{Data Construction Pipeline (Figure~\ref{fig:framework})}

\paragraph{Scratchblocks Generation.}

We begin from the public Scratch project corpus of \citet{aivaloglou2017scratch}. Scratch projects are internally represented as JSON-like graphs of interconnected blocks rather than linear source code. We therefore convert executable scripts into normalized \textit{scratchblocks}~\cite{scratchblocks2024}, a textual notation for representing Scratch blocks. This conversion preserves event triggers, block ordering, control-flow hierarchy, and slot arguments while providing a compact, human-readable representation suitable for language modeling. 
We use \textit{scratchblocks} as the generation target rather than the native Scratch project format. Compared with the original JSON-based representation, \textit{scratchblocks} is substantially more compact and structurally explicit while remaining faithful to the underlying program structure. It also enables parser-based validation and semantic analysis without requiring implementation-specific metadata contained in raw project files.

To ensure structural validity, we validate every extracted program using the Scratch VM parser and discard scripts that fail to parse. We further remove exact duplicates based on their normalized \textit{scratchblocks} representations. The resulting corpus consists exclusively of parser-valid Scratch programs.

\paragraph{Natural Language Generation.}

For each \emph{scratchblock} program, we generate the natural-language description in two stages. 
First, we generate a deterministic description from the \emph{scratchblock} using a rule-based converter that preserves program structure, control flow, numeric arguments, and named entities. The resulting descriptions are tightly aligned with the underlying program semantics but often exhibit rigid and repetitive phrasing. We therefore rewrite these descriptions using an LLM under strict semantic-preservation constraints. The model is instructed to improve fluency and student-like naturalness, while requiring all executable semantics (e.g., events, actions, conditions, numeric arguments, and named entities) to remain unchanged.

This two-stage pipeline separates semantic grounding from stylistic naturalization: program semantics originate entirely from the executable \emph{scratchblock}, while the rewriting stage improves linguistic quality and pedagogical realism without modifying executable behavior.

\paragraph{Semantic Alignment Consistency}

\label{sec:sac}

Although parser validation guarantees \emph{structural} correctness of the program side, it does not ensure that the rewritten description remains faithful to the underlying program. Thus, we introduce SAC, a scalable slot-level verification procedure for measuring semantic agreement between natural-language descriptions and Scratch programs.

Given an NL--program pair $(x,p)$, SAC extracts semantic slots from both representations, including event types, actions, control-flow constructs, numeric arguments, and named entities such as variables, costumes, backdrops, and broadcast messages (Table~\ref{tab:sac_slots}). Agreement is computed independently for each slot. Scalar attributes are evaluated by exact match, whereas set-valued attributes are compared using Jaccard overlap:
$
\mathrm{score}(A,B)=
{|A\cap B|}/{|A\cup B|}.
\label{eq:jaccard_score}
$
where $A$ and $B$ are the slot-value sets extracted from scratchblocks $p$ and natural language $x$, respectively. 
The final SAC score is obtained by averaging over all comparable non-empty slots:
\begin{equation}
\mathrm{SAC}(x,p)
=
\frac{1}{|\mathcal{S}_{x,p}|}
\sum_{s\in\mathcal{S}_{x,p}}
\mathrm{score}_s(x,p)
\label{eq:agg}
\end{equation}
where $\mathcal{S}_{x,p}$ denotes the set of comparable slots. We define a pair as \emph{high-confidence aligned} if $\mathrm{SAC}(x, p) \geq 0.85$, a threshold chosen to balance semantic precision and evaluation-pool coverage.
SAC serves two purposes: (i) it enables scalable filtering of semantically inconsistent examples introduced during language rewriting, and (ii) it provides fine-grained slot-level diagnostics that support detailed analysis of model errors beyond overall correctness metrics.

\subsection{Diagnostic Benchmark Construction}
\label{sec:benchmark}

In addition to the full train/validation/test splits, we construct a controlled diagnostic benchmark for rigorous evaluation and analysis. This set aims to provide a semantically cleaner and more structurally informative evaluation set than the raw test distribution while preserving coverage over the core phenomena of Scratch programming.

Starting from the canonical test split with 31,166 samples, we apply SAC and retain only examples satisfying $\mathrm{SAC}\ge0.85$, yielding a high-confidence pool of 23,594 semantically aligned instances (75.6\% of the original test split; Figure~\ref{fig:sac_distribution}). This filtering removes examples where the rewritten NL description may have drifted from the underlying executable behavior despite remaining fluent and grammatically correct.

From this pool, we construct the final 800-example diagnostic benchmark by maximizing coverage over Scratch-specific semantic attributes. Specifically, we ensure diversity across attributes, including event types, action types, loop types, condition types, and the presence of key semantic attributes such as numeric arguments, variable references, costume names, backdrop names, and broadcast messages. To avoid over-representing only the simplest programs and ensure broader block-based behaviors, selection is performed to encourage diversity across these structural dimensions.

\definecolor{hj_bl}{HTML}{2171B5}
\definecolor{hj_pi}{HTML}{F768A1}

\begin{figure}
    \centering
    \includegraphics[width=0.9\linewidth]{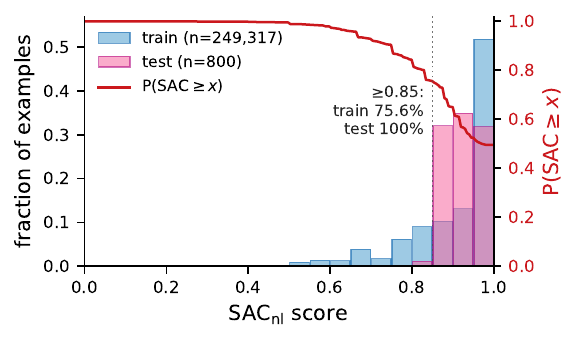}
    \caption{SAC$_{\mathrm{nl}}$ distribution: training set (\textcolor{hj_bl}{blue}) vs.\ filtered benchmark (\textcolor{hj_pi}{pink}, {SAC} $\geq 0.85$).}
\label{fig:sac_distribution}
\end{figure}

\begin{figure}[t]
    \centering
    \includegraphics[width=0.95\linewidth]{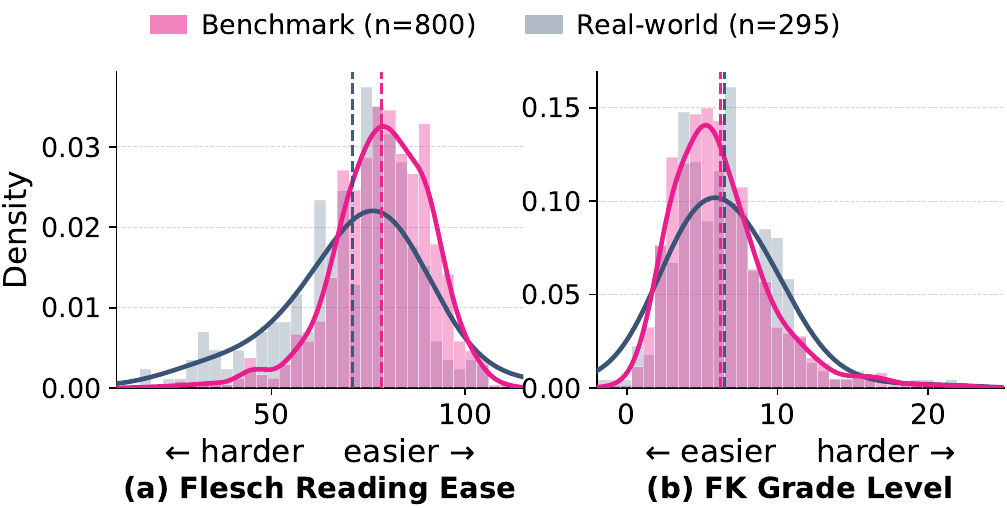}
    \caption{Distributions of Flesch Reading Ease and FK Grade Level for benchmark descriptions ($n{=}800$) and real-world learner prompts ($n{=}295$).
}
    \label{fig:readability}
\end{figure}

As a secondary balancing signal, we also consider the readability of the rewritten NL descriptions. For each description, we compute two standard readability measures: Flesch Reading Ease (higher values indicate greater readability) and Flesch--Kincaid Grade Level (higher values indicate greater syntactic and lexical complexity). We convert both measures into standardized $z$-scores over the retained high-confidence pool and combine them into a single readability score:
\begin{equation}
q_{\mathrm{NL}}(x) = z\!\left(\mathrm{Flesch}(x)\right) - z\!\left(\mathrm{FKGrade}(x)\right),
\end{equation}
where higher values correspond to descriptions that are easier to read relative to the rest of the pool. This score is not used as a hard filtering criterion; instead, after structural-semantic coverage has been prioritized, it serves as a soft balancing signal to encourage diversity in linguistic complexity. To assess whether the generated NL descriptions resemble realistic learner inputs, we additionally compare their readability distribution against real-world learner prompts from \citet{si2026exploring}. As shown in Figure~\ref{fig:readability}, the two distributions largely overlap under both measures, suggesting that the benchmark captures a similar range of linguistic complexity to authentic learner queries.

\begin{table}[t]
\centering
\scalebox{0.75}{
\begin{tabular}{l >{\raggedright\arraybackslash}p{6.3cm}}
\toprule
\textbf{Slot} & \textbf{Captures} \\
\midrule
\texttt{event\_type}       & Kind of triggering event (flag, key, message, click) \\
\texttt{event\_target}     & Specific entity triggering the event \\
\texttt{action\_types}     & Set of action verbs performed \\
\texttt{loop\_types}       & Kind(s) of loop construct \\
\texttt{loop\_counts}      & Iteration counts for bounded loops \\
\texttt{condition\_types}  & Predicate types appearing in conditionals \\
\texttt{numeric\_values}   & Numeric literals in block slots \\
\texttt{variable\_names}   & Variable identifiers referenced \\
\texttt{costume\_names}    & Costume identifiers referenced \\
\texttt{backdrop\_names}   & Backdrop identifiers referenced \\
\texttt{broadcast\_names}  & Broadcast message identifiers \\
\bottomrule
\end{tabular}
}
\caption{The eleven behavioral slots extracted for SAC. For each $(x, p)$ pair, slots are populated structurally 
from the scratchblocks and through pattern-based extractors from NL; agreement is computed slot-wise.}
\label{tab:sac_slots}
\end{table}

The resulting 800-example subset serves as the main evaluation benchmark used in our experiments. Compared with the raw test split, it offers three advantages: (i) stronger semantic reliability through SAC-based filtering, (ii) a more controlled basis for fine-grained comparison of model behavior, and (iii) broader structural coverage over Scratch-specific program phenomena.

%% file: Contents/04_experiments.tex
\section{Experiments}

\paragraph{Models.}

We evaluate both proprietary and open-weight LLMs. Proprietary models include GPT-4.1, GPT-4o-mini, and a fine-tuned GPT-4o-mini. Open-weight baselines include FLAN-T5~\cite{chung2024scaling}, Qwen2.5-7B-Instruct~\cite{qwen2025qwen25}, Llama-3.1-8B-Instruct, together with their fine-tuned variants. 

\paragraph{Metrics.}

We report both conventional generation metrics and SAC. Conventional metrics include exact match (EM), token-level F1, and parse success (PS), which measure exact program agreement, token overlap, and parser validity, respectively. Although these metrics capture lexical similarity and syntactic validity of the program, they do not directly assess whether a generated program is semantically aligned with either the input description or the reference program. We therefore additionally report SAC, a slot-level measure of semantic agreement. SAC$_{\text{nl}}$ evaluates alignment between a generated program and its input natural-language description, whereas SAC$_{\text{ref}}$ evaluates alignment with the reference program. For both variants, we report the average score (Avg), the proportion of perfectly aligned examples (\(=100\%\)), and the proportion of highly aligned examples (\(\geq85\%\)).

\paragraph{Training and Inference.}

For proprietary models, we evaluate GPT-4.1 and GPT-4o-mini using retrieval-based in-context learning (ICL), following prior work on domain-specific language generation~\citep{10.1145/3654992}. For each test instance, we retrieve the 20 most similar training examples using TF--IDF similarity and include them as demonstrations together with task instructions and constraints. Inference is performed with OpenAI Batch API, and detailed prompts are provided in Appendix~\ref{appen:prompt}. For supervised fine-tuning (SFT), we fine-tune GPT-4o-mini for three epochs using 800 training examples and 100 validation examples selected to maximize coverage of event types and program structures. Due to the cost of proprietary-model fine-tuning, we use this representative subset rather than the full training split.
For open-weight models, we fine-tune FLAN-T5 and Qwen2.5-7B-Instruct on the constructed NL2Scratch training set. During evaluation, all models directly generate scratchblocks, which is subsequently normalized and validated using the Scratch parser.

\paragraph{SAC-based Reranking.}
\label{sec:reranking}
Beyond evaluation, we investigate whether SAC can serve as an effective inference-time selection signal. Given an input NL description $x$, we first sample a set of candidate programs $\{\hat P_1,\ldots,\hat P_N\}$ and then rerank them according to their semantic alignment with the input. The final prediction is selected as:
\[
\hat P^\star
=
\arg\max_i
\mathrm{SAC}_{nl}(\hat P_i,x).
\]
As SAC directly measures semantic agreement between the generated program and the intended description, this procedure favors semantically faithful candidates rather than those with merely high model likelihood. 
Further implementation details are provided in Appendix~\ref{app:training_details}.

%% file: Contents/05_results_hj.tex
\newcommand{\gc}{\cellcolor{hj_gray!20}}
\newcommand{\gco}{\cellcolor{hj_gray!50}}

\begin{table*}
\centering
\scalebox{0.72}{
\begin{tabular}{l l ccc c ccc c ccc}
\toprule
& & & & & & \multicolumn{3}{c}{SAC$_{\text{nl}}$} & & \multicolumn{3}{c}{SAC$_{\text{ref}}$} \\
\cmidrule(lr){7-9} \cmidrule(lr){11-13}
& \textbf{Model} & EM & F1 & PS & & Avg & =100\% & $\geq$85\% & & Avg & =100\% & $\geq$85\% \\
\midrule
\multirow{5}{*}{\rotatebox[origin=c]{90}{\textit{Closed-sourced}}}
&{GPT-4.1}
& 3.4 {\scriptsize $\pm$1.2}
& 83.1 {\scriptsize $\pm$0.7}
& 52.8 {\scriptsize $\pm$3.5}
& & 89.3 {\scriptsize $\pm$0.7}
& 12.6 {\scriptsize $\pm$2.3}
& 82.1 {\scriptsize $\pm$2.7}
& & 93.9 {\scriptsize $\pm$0.7}
& 51.7 {\scriptsize $\pm$3.4}
& 84.0 {\scriptsize $\pm$2.6} \\

& \gc {+ few-shot (20)}
& \gc \textbf{\gc 67.8 {\scriptsize $\pm$3.3}}
& \gc \textbf{97.1 {\scriptsize $\pm$0.4}}
& \gc 97.6 {\scriptsize $\pm$1.1}
& \gc & \gc 92.7 {\scriptsize $\pm$0.4}
& \gc \textbf{18.9 {\scriptsize $\pm$2.7}}
& \gc \textbf{98.9 {\scriptsize $\pm$0.8}}
& \gc & \gc \textbf{99.1 {\scriptsize $\pm$0.2}}
& \gc \textbf{88.2 {\scriptsize $\pm$2.3}}
& \gc \textbf{99.1 {\scriptsize $\pm$0.7}} \\

&{GPT-4o-mini}
& 2.6 {\scriptsize $\pm$1.1}
& 77.9 {\scriptsize $\pm$1.0}
& 22.0 {\scriptsize $\pm$2.9}
& & 75.8 {\scriptsize $\pm$1.4}
& 10.2 {\scriptsize $\pm$2.2}
& 42.0 {\scriptsize $\pm$3.4}
& & 78.0 {\scriptsize $\pm$1.6}
& 25.1 {\scriptsize $\pm$3.0}
& 42.5 {\scriptsize $\pm$3.4} \\

 & \gc {+ few-shot (20)}
& \gc 62.1 {\scriptsize $\pm$3.4}
& \gc 96.2 {\scriptsize $\pm$0.5}
& \gc 93.4 {\scriptsize $\pm$1.8}
& \gc & \gc 91.7 {\scriptsize $\pm$0.5}
& \gc 18.4 {\scriptsize $\pm$2.7}
& \gc 94.4 {\scriptsize $\pm$1.6}
& \gc & \gc 97.8 {\scriptsize $\pm$0.5}
& \gc 82.5 {\scriptsize $\pm$2.6}
& \gc 95.9 {\scriptsize $\pm$1.5} \\

 & \gco {+ SFT}
& \gco 53.8 {\scriptsize $\pm$3.6}
& \gco 95.3 {\scriptsize $\pm$0.6}
& \gco \textbf{98.5 {\scriptsize $\pm$0.9}}
& \gco & \gco 92.0 {\scriptsize $\pm$0.4}
& \gco 17.9 {\scriptsize $\pm$2.8}
& \gco 94.4 {\scriptsize $\pm$1.6}
& \gco & \gco 97.7 {\scriptsize $\pm$0.4}
& \gco 75.6 {\scriptsize $\pm$3.0}
& \gco 95.6 {\scriptsize $\pm$1.4} \\

\midrule
\multirow{6}{*}{\rotatebox[origin=c]{90}{\textit{Open-sourced}}}
&FLAN-T5
& 27.5 {\scriptsize $\pm$3.1}
& 92.9 {\scriptsize $\pm$0.6}
& 71.5 {\scriptsize $\pm$3.1}
& & 91.6 {\scriptsize $\pm$0.5}
& 18.0 {\scriptsize $\pm$2.6}
& 91.1 {\scriptsize $\pm$2.0}
& & 90.2 {\scriptsize $\pm$0.7}
& 38.2 {\scriptsize $\pm$3.4}
& 64.5 {\scriptsize $\pm$3.3} \\

& \gco \quad + SAC-based rerank
& \gco 27.5 {\scriptsize $\pm$3.0}
& \gco 92.7 {\scriptsize $\pm$0.6}
& \gco 71.1 {\scriptsize $\pm$3.1}
& \gco & \gco 91.7 {\scriptsize $\pm$0.4}
& \gco 18.6 {\scriptsize $\pm$2.7}
& \gco 91.6 {\scriptsize $\pm$1.9}
& \gco & \gco 96.7 {\scriptsize $\pm$0.5}
& \gco 70.8 {\scriptsize $\pm$3.1}
& \gco 93.6 {\scriptsize $\pm$1.7} \\

&Llama-3.1-8B-Instruct
& 24.1 {\scriptsize $\pm$3.0}
& 80.8 {\scriptsize $\pm$1.6}
& 81.0 {\scriptsize $\pm$2.7}
& & 90.6 {\scriptsize $\pm$0.5}
& 14.9 {\scriptsize $\pm$2.5}
& 87.6 {\scriptsize $\pm$2.3}
& & 96.5 {\scriptsize $\pm$0.4}
& 66.1 {\scriptsize $\pm$3.3}
& 94.1 {\scriptsize $\pm$1.7} \\
& \gco \quad + SAC-based rerank
& \gco 26.8 {\scriptsize $\pm$3.1}
& \gco 81.5 {\scriptsize $\pm$1.6}
& \gco 80.0 {\scriptsize $\pm$2.8}
& \gco & \gco \textbf{92.8 {\scriptsize $\pm$0.4}}
& \gco \textbf{18.9 {\scriptsize $\pm$2.7}}
& \gco 98.5 {\scriptsize $\pm$0.8}
& \gco & \gco 98.7 {\scriptsize $\pm$0.2}
& \gco 81.6 {\scriptsize $\pm$2.7}
& \gco 98.9 {\scriptsize $\pm$0.7} \\
&Qwen2.5-7B-Instruct
& 61.6 {\scriptsize $\pm$3.4}
& 94.6 {\scriptsize $\pm$0.7}
& 96.8 {\scriptsize $\pm$1.2}
& & 91.7 {\scriptsize $\pm$0.4}
& 17.1 {\scriptsize $\pm$2.6}
& 93.2 {\scriptsize $\pm$1.8}
& & 97.9 {\scriptsize $\pm$0.4}
& 80.0 {\scriptsize $\pm$2.8}
& 95.6 {\scriptsize $\pm$1.4} \\
& \gco \quad + SAC-based rerank
& \gco 61.6 {\scriptsize $\pm$3.4}
& \gco 96.1 {\scriptsize $\pm$0.5}
& \gco 98.0 {\scriptsize $\pm$0.9}
& \gco & \gco 92.7 {\scriptsize $\pm$0.3}
& \gco 18.4 {\scriptsize $\pm$2.7}
& \gco 98.8 {\scriptsize $\pm$0.8}
& \gco & \gco 98.9 {\scriptsize $\pm$0.2}
& \gco 85.5 {\scriptsize $\pm$2.4}
& \gco 98.8 {\scriptsize $\pm$0.8} \\
\bottomrule
\end{tabular}
}
\caption{Model performance comparison. EM and F1 measures token-level overlap between generated ($\hat{P}$) and gold ($P$) program, and parse success (PS) rate denotes the percentage of generations accepted by the Scratch parser. SAC$_{nl}$ aligns $\hat{P}$ with NL, SAC$_{\text{ref}}$ aligns $\hat{P}$ with $P$. For each SAC, we report the mean score (Avg) and the proportion of samples with perfect (=100\%) and high ($\geq$85\%) alignment. Small $\pm$ values denote 95\% bootstrap confidence intervals computed over evaluation examples. Best results per column in \textbf{bold}.}
\label{tab:results_800_full}
\end{table*}

\section{Results}

Our empirical analysis is organized around three questions:
(§~\ref{res:overall}) How well do current LLMs generate Scratch programs from natural-language descriptions?
(§~\ref{res:metric}) To what extent do conventional metrics reflect semantic faithfulness?
(§~\ref{res:error}) Which semantic components are most error-prone in generated programs?

\subsection{Overall Generation Performance}\label{res:overall}

Table~\ref{tab:results_800_full} reports generation performance on the NL2Scratch diagnostic benchmark. Among proprietary models, GPT-4.1 with few-shot prompting achieves the strongest overall results. GPT-4o-mini shows similar trend under few-shot prompting, while SFT substantially improves performance relative to zero-shot generation. Among open-weight models, fine-tuned Qwen2.5-7B-Instruct performs best, achieving 61.6 EM, 94.6 F1, and 96.8 parse success, considerably outperforming FLAN-T5 and Llama-3.1-8B-Instruct.

Across model families, few-shot prompting provides large gains over zero-shot inference, highlighting the importance of structural demonstrations for Scratch generation. SFT and SAC-based reranking further improves exact match and parser validity. However, despite near-perfect F1 and PS rates, perfect semantic alignment remains substantially lower across all models. This indicates that high token-level similarity and syntactic validity do not necessarily translate into behaviorally faithful programs. Overall, current LLMs are capable of generating parser-valid Scratch programs at high rates, but faithfully reproducing the intended program semantics remains challenging.

\subsection{Lexical Similarity Overestimates Semantic Alignment}\label{res:metric}

A central finding of our study is that strong performance under conventional metrics does not necessarily imply semantically matched Scratch programs. While most models achieve very high token-level F1 scores, substantially lower performance is observed under stricter semantic criteria.

For example, GPT-4.1 with few-shot prompting achieves 97.1 F1, yet only 18.9\% of predictions achieve perfect SAC$_{nl}$ alignment. Similarly, Qwen2.5-7B-Instruct obtains 94.6 F1 but only 17.1\% perfect SAC$_{nl}$ alignment. Even when generated programs closely resemble the reference at the token level, they often differ in important slot behavior details such as event triggers, conditions, actions, or numeric arguments. These discrepancies often remain invisible to overlap-based metrics or grammar-grounded execution success rate.

The same pattern is reflected in SAC$_{ref}$. Average semantic alignment scores are generally high, indicating that models recover much of the intended program content. However, exact semantic agreement remains substantially lower than token-level similarity, suggesting that models often reconstruct the broad structure of a program without faithfully reproducing all executable details.
These results demonstrate that conventional overlap-based metrics notably overestimate slot-level semantic behavioral agreement in block-based program generation, suggesting reliable evaluation requires explicit verification of program semantics.

\begin{figure}[t]
    \centering
    \includegraphics[width=0.9\columnwidth]{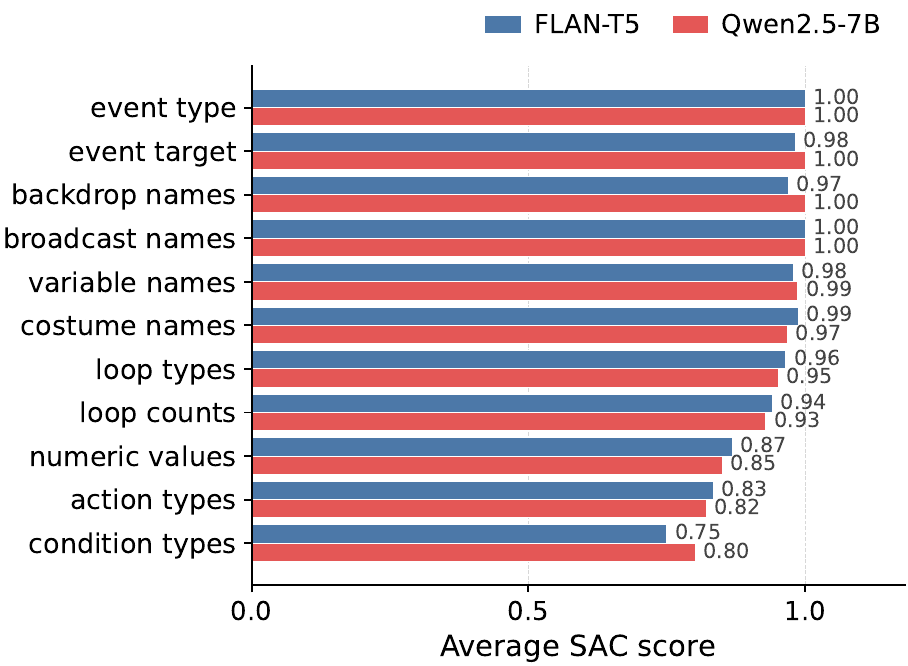}
    \caption{Average SAC$_{nl}$ score by semantic slot.
    }
    \label{fig:slot_avg_sac}
\end{figure}
% \vspace{-4pt}

\begin{figure*}[t]
    \centering
    \includegraphics[width=0.9\linewidth]{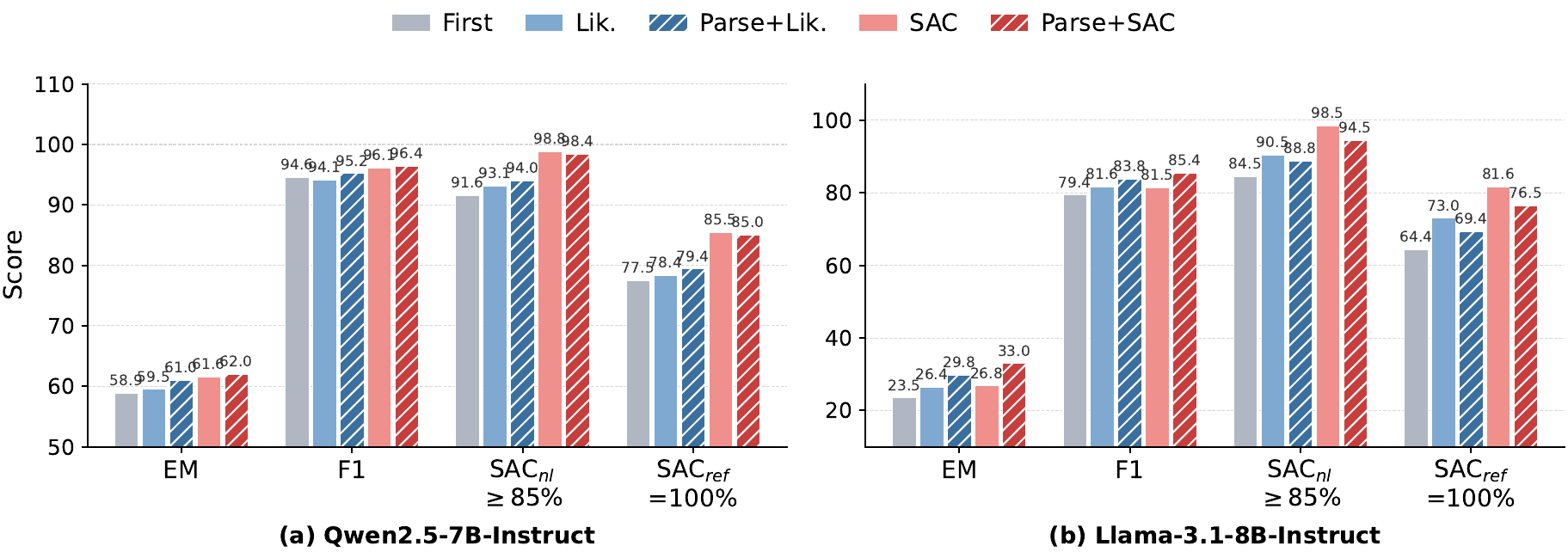}
    \caption{\textbf{Effect of reranking strategies on generation quality.} We compare first-candidate decoding, likelihood-based reranking (Lik.), parser-aware reranking (Parse+Lik.), SAC-based reranking (SAC), and parser-constrained SAC reranking (Parse+SAC) for Qwen2.5-7B-Instruct and Llama-3.1-8B-Instruct. While likelihood- and parser-based reranking yield modest improvements, SAC-guided reranking consistently produces the strongest gains in semantic metrics, notably increasing SAC$_{\text{nl}}$ and SAC$_{\text{ref}}$ alignment. These results suggest that explicit semantic verification provides a more effective reranking signal than lexical likelihood or parser validity alone.
}
    \label{fig:reranking}
\end{figure*}

\subsection{Where Do Models Fail?}\label{res:error}

To better understand model behavior beyond aggregate metrics, we analyze performance at the level of individual SAC slots. Figure~\ref{fig:slot_avg_sac} reports average slot-level alignment scores for FLAN-T5 and Qwen2.5-7B-Instruct. The results reveal a clear distinction between structural scaffolding and operational details. Event-related and entity-centric slots are recovered almost perfectly across models. Event types, broadcast names, backdrop names, variable names, and costume names all achieve near-ceiling alignment scores, indicating that models reliably capture the overall context and referenced program entities.

In contrast, the most challenging slots are those that directly determine program behavior. Condition types, action types, and numeric values consistently obtain the lowest alignment scores across models. For example, condition-type agreement remains substantially lower than event-related slots even for the strongest model, while action and numeric slots exhibit persistent errors. These findings indicate that current models generally recover the high-level structure of Scratch programs but struggle with the operational details required for exact behavioral reproduction. Importantly, these are precisely the errors that remain largely hidden under conventional overlap-based metrics, despite having a direct impact on program execution.

\section{Discussions}\label{sec:discussion}

\paragraph{The Gap Widens with Program Complexity.}

\begin{figure}
    \centering
    \includegraphics[width=\linewidth]{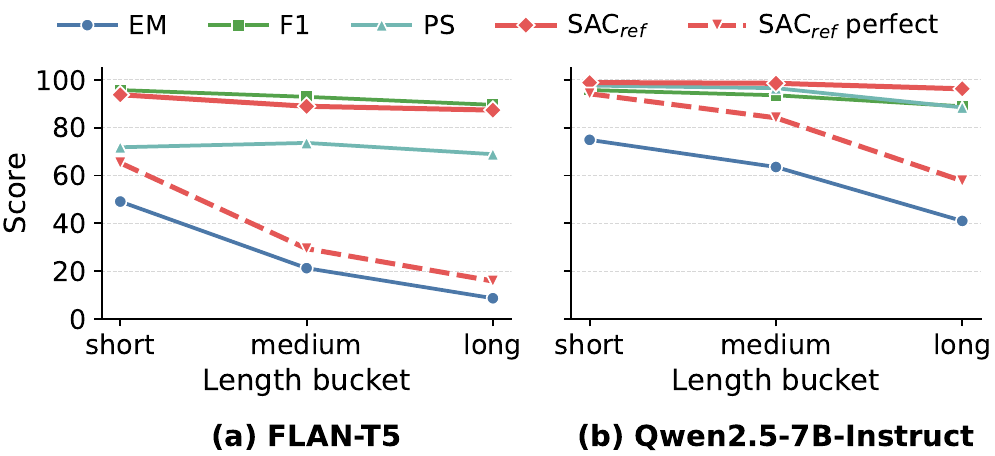}
    \caption{Per-bucket performance by program length. As targets become longer,
    strict metrics (EM, SAC$_{\text{ref}}${\small-perfect}) degrade much more sharply than lenient ones.}
    \label{fig:length}
\end{figure}

The discrepancy between lexical similarity and semantic alignment becomes more pronounced as programs grow longer. Figure~\ref{fig:length} shows that exact match and perfect semantic alignment deteriorate substantially with increasing program length, whereas token-level F1 remains comparatively stable. For FLAN-T5, exact match drops from 49.1\% on short programs to only 8.8\% on long programs, while F1 decreases only modestly. Qwen2.5-7B-Instruct exhibits the same trend: exact match falls from 74.9\% to 41.0\%, and perfect SAC$_{ref}$ alignment declines sharply, despite F1 remaining close to 90\%. These results suggest that errors accumulate as the number of interacting blocks, arguments, and dependencies increases. This behavior highlights a key limitation of overlap-based evaluation. As programs become more complex, local token recovery remains relatively easy, whereas preserving global executable structure becomes more difficult. Consequently, surface similarity becomes progressively less informative as a proxy for correctness.

\paragraph{Can Semantic Verification Improve Decoding?}

Beyond evaluation, semantic verification can also improve generation quality. Figure~\ref{fig:reranking} compares several reranking strategies for open-weight models, including likelihood-based ranking, parser-aware ranking, and SAC-guided reranking. Across both Qwen2.5-7B-Instruct and Llama-3.1-8B-Instruct, SAC-guided reranking consistently yields the strongest semantic performance. For example, Qwen improves from 80.0\% to 85.5\% perfect SAC$_{ref}$ alignment, while Llama improves from 66.1\% to 81.6\%. Similar gains are observed for high-confidence semantic alignment rates. These results suggest that semantic verification is useful not only for benchmarking, but also as a decoding signal. Explicit semantic feedback can help distinguish superficially plausible programs from behaviorally faithful ones, providing a promising direction for future structure-aware generation.

%% file: Contents/08_Conclusion.tex
\section{Conclusion}

We introduced NL2Scratch, the first executable benchmark for natural-language-to-Scratch generation, together with Semantic Alignment Consistency (SAC), a scalable slot-level evaluation framework for semantic agreement between NL descriptions and Scratch programs. Across proprietary and open-weight LLMs, we found that strong performance under conventional metrics frequently masks slot-level semantic errors, with failures concentrated in actions, conditions, and numeric arguments and becoming more pronounced as program complexity increases. These findings suggest that block-based programming constitutes a distinct NL2Code regime where lexical overlap and parser validity alone are insufficient indicators of accuracy, motivating future work on structure-aware generation and evaluation methods.

%% file: Contents/07_Limitations.tex
\section{Limitations}
While NL2Scratch is designed to promote semantic alignment between natural-language descriptions and executable programs through parser validation, LLM-based rewriting, and SAC-based verification, we do not explicitly evaluate the linguistic realism of the generated descriptions. Although the rewriting stage aims to produce student-like language, it remains unclear to what extent the resulting descriptions reflect the vocabulary, ambiguity, and instruction styles used by real learners. Future work could validate this aspect using authentic learner-authored descriptions or direct comparisons with naturally occurring learner instructions. 

Second, SAC is designed as an interpretable and scalable semantic verification framework rather than a complete behavioral equivalence checker. Although it captures a broad range of semantically important program attributes, including events, actions, control-flow structures, and arguments, programs with similar SAC scores may still differ in subtle runtime behaviors arising from complex interactions across scripts, state-dependent execution, or event synchronization. Extending semantic verification to richer execution-level behavioral analysis remains an important direction for future work. 

Finally, our study focuses exclusively on Scratch, a representative block-based programming environment. While many of the challenges identified in this work stem from characteristics common to block-based programming, such as event-driven execution, visual composition, and distributed program structure, it remains unclear whether the same observations generalize to other educational programming platforms or visual programming paradigms. Evaluating the proposed benchmarking methodology across additional environments would help establish the broader applicability of our findings.

%% file: Contents/08_ethical.tex
\section*{Ethical Statement}
NL2Scratch is intended to support research on natural-language-to-program generation in block-based programming environments commonly used in early programming education. The benchmark is constructed from publicly available Scratch projects, and all generated natural-language descriptions are derived from program content, without including any personal user information. AI assistance was used for language editing and proofreading.

\section*{Acknowledgments}
This project was supported by ETH AI Center Postdoctoral Fellowships to Heejin Do. Also, this research was supported by the Swiss AI Initiative, and the Swiss AI Large Grant SCR1089274.

%% file: Contents/09_appendix.tex
\section{Prompts}
\label{appen:prompt}

\begin{promptbox}
{NL2Scratch Prompt used for GPT}
\small
\role{Task} Convert natural language descriptions into Scratch pseudocode.
\smallskip\smallskip

\role{Scratch Pseudocode Syntax} Scratch pseudocode syntax rules and requirements:\\
\smallskip
1. Output only Scratch pseudocode, with one block per line and no markdown.\\
2. Use exact Scratch-style block text from the examples. Do not output opcode keys.\\
3. Stack/command blocks are plain lines, e.g. \texttt{move (10) steps}.\\
4. Reporter inputs must be wrapped in parentheses, e.g. \texttt{(x position)}, \texttt{((score) + (1))}.\\
5. Boolean conditions must be wrapped in angle brackets, e.g. \texttt{<mouse down?>}, \texttt{<touching (edge v)?>}.\\
6. Text/name inputs use square brackets, e.g. \texttt{[message1]}, \texttt{[costume1]}, \texttt{[score v]}.\\
7. Menu/dropdown inputs include the v marker, e.g. \texttt{(space v)}, \texttt{(random position v)}, \texttt{[all v]}.\\
8. Control blocks use exact forms such as \texttt{repeat (10)}, \texttt{forever}, \texttt{if <condition> then}, \texttt{if <condition> then else}, \texttt{wait until <condition>}.\\
9. Indent nested blocks with exactly 4 spaces and close every C-block with \texttt{end}.\\
10. For if else, use: \texttt{if <condition> then}, true branch, \texttt{else}, false branch, \texttt{end}.\\
11. Preserve names, messages, numbers, signs, decimal values, and action order from the natural language.\\
12. If no event is described, do not invent one.
\smallskip\smallskip

\role{Retrieved Examples} Here are some examples: \\
20 * Example (Natural Language, Pseudocode)

\noindent\rule{\linewidth}{0.4pt}
% \vspace{0.6em}
\role{Test Item} Now, please generate the Scratch pseudocode for the following description: \\
\textbf{Natural Language:} \texttt{\{{natrual\_language\_query}\}} \\
\textbf{Pseudocode:} \textcolor{purple}{[To be generated]}
\end{promptbox}

\begin{promptbox}
{Natural-Language Rewriting Prompt used for Dataset Construction}
\small
\role{System Prompt}
You are helping build a dataset for kids learning Scratch. Rewrite the base instruction so it sounds like a natural instruction a kid would type to the machine.

\smallskip
\role{Rules}
\begin{enumerate}
    \item Keep the meaning exactly the same.
    \item Do not add or remove any steps.
    \item Keep all numbers exactly the same.
    \item Keep triggers and conditions explicit.
    \item Keep variable, sprite, costume, backdrop, and message names unchanged.
    \item Remove awkward menu markers and quoting when possible.
    \item For key names, say things like ``q key'' instead of ``q v''.
    \item Do not output pseudocode.
    \item Output only the rewritten natural-language instruction.
\end{enumerate}

\noindent\rule{\linewidth}{0.4pt}

\role{User Prompt Template}
\textbf{Scratchblocks pseudocode:}\\
\texttt{\{pseudocode\}}

\smallskip
\textbf{Base instruction:}\\
\texttt{\{base\_nl\}}

\smallskip
Rewrite it as one natural child-like instruction:
\end{promptbox}

\section{Additional Dataset Statistics}
\label{app:dataset_stats}

Table~\ref{tab:sac_slot_coverage_subsets} reports SAC-slot coverage across the full parser-valid corpus, the SAC-filtered high-confidence pool, and the final 800-example diagnostic benchmark. Compared with the full corpus, the diagnostic benchmark increases coverage of structurally informative phenomena such as loops, conditions, variables, backdrops, and broadcasts, reflecting its role as a controlled evaluation set rather than a distribution-matched sample.

\begin{table}[h]
\centering
\scalebox{0.8}{
\begin{tabular}{lrrr}
\toprule
Slot & Full & Pool & Diag. \\
\midrule
\texttt{action\_types} & 100.0 & 100.0 & 100.0 \\
\texttt{loop\_types} & 41.9 & 41.5 & 67.9 \\
\texttt{condition\_types} & 30.1 & 29.8 & 61.3 \\
\texttt{numeric\_values} & 92.6 & 92.7 & 90.4 \\
\texttt{variable\_names} & 30.3 & 30.1 & 49.8 \\
\texttt{costume\_names} & 20.7 & 20.3 & 29.0 \\
\texttt{backdrop\_names} & 13.5 & 13.6 & 20.0 \\
\texttt{broadcast\_names} & 0.3 & 0.2 & 5.4 \\
\bottomrule
\end{tabular}
}
\caption{SAC slot coverage across dataset subsets. Values are percentages of examples whose gold pseudocode contains at least one extracted value for the slot. Full denotes the full parser-valid corpus, Pool the SAC-filtered high-confidence test pool, and Diag. the 800-example diagnostic benchmark.}
\label{tab:sac_slot_coverage_subsets}
\end{table}

Table~\ref{tab:dataset_funnel} summarizes the dataset construction pipeline. Starting from 311,648 parser-valid NL--pseudocode pairs, SAC filtering yields a high-confidence pool of 23,594 examples, from which the final 800-example diagnostic benchmark is sampled. Together, these statistics show that NL2Scratch provides both large-scale coverage through the full corpus and a compact, structurally enriched diagnostic benchmark for controlled model evaluation.

\begin{table}[h]
\centering
\small
\begin{tabular}{lr}
\toprule
Stage & Count \\
\midrule
Full parser-valid NL--pseudocode corpus & 311,648 \\
Test split & 31,166 \\
SAC-filtered high-confidence test pool & 23,594 \\
Diagnostic benchmark & 800 \\
\bottomrule
\end{tabular}
\caption{Dataset sizes at the main construction and filtering stages.}
\label{tab:dataset_funnel}
\end{table}

\section{SAC Implementation Details}
\label{app:sac_details}

Semantic Alignment Consistency (SAC) is implemented as a rule-based slot-matching procedure that compares a Scratchblocks program against a natural-language description. SAC is not intended to decide full program equivalence; instead, it provides a reproducible and interpretable semantic consistency signal over behaviorally important components of Scratch programs, including event structure, control flow, numeric arguments, and named entities. The main implementation is provided in the released code under the semantic-alignment module.

\paragraph{Overview.}
For each pair $(x,p)$, consisting of a natural-language description $x$ and a Scratch pseudocode program $p$, SAC performs three steps:
\begin{enumerate}
    \item extract a structured slot representation from the pseudocode;
    \item extract a corresponding slot representation from the natural language;
    \item compare the extracted slots with slot-specific similarity functions.
\end{enumerate}
The output includes per-slot scores, an overall alignment score, a binary perfect-alignment flag, a binary high-confidence-alignment flag, and a list of mismatched slots. We define perfect alignment as $\mathrm{SAC}=1.0$ and high-confidence alignment as $\mathrm{SAC}\geq0.85$.

\paragraph{Slot schema.}
SAC uses the ordered slot set from Table~\ref{tab:sac_slots}.
These slots cover the main semantic components of Scratch programs: triggering events, program actions, loop and condition structure, numeric arguments, and references to variables or multimodal assets.

\paragraph{Pseudocode-side extraction.}
Slot extraction from scratchblocks pseudocode is largely deterministic because the representation follows a normalized block syntax. SAC first identifies the event hat block at the beginning of the script. The extractor recognizes the following coarse event categories:
\begin{itemize}
    \item \texttt{when @greenFlag clicked} $\rightarrow$ \texttt{green\_flag};
    \item \texttt{when backdrop switches to [\dots]} $\rightarrow$ \texttt{backdrop\_switch};
    \item \texttt{when I receive [\dots]} $\rightarrow$ \texttt{broadcast\_receive};
    \item \texttt{when [key] key pressed} $\rightarrow$ \texttt{key\_press};
    \item \texttt{when this sprite clicked} $\rightarrow$ \texttt{sprite\_clicked}.
\end{itemize}
When applicable, the corresponding target is also extracted into \texttt{event\_target} and inserted into the relevant name slot, such as \texttt{broadcast\_names} or \texttt{backdrop\_names}.

Action types are extracted through explicit line-prefix matching. For example, lines beginning with \texttt{show}, \texttt{hide}, \texttt{move}, \texttt{glide}, \texttt{turn left}, \texttt{turn right}, \texttt{point in direction}, \texttt{say}, \texttt{think}, \texttt{wait}, \texttt{broadcast}, \texttt{play sound}, \texttt{stop all sounds}, and \texttt{ask} are mapped to canonical action labels in \texttt{action\_types}. Costume and backdrop changes are also recorded in the corresponding named-entity slots.

Loop structure is extracted directly from Scratch control syntax. A line equal to \texttt{forever} adds \texttt{forever} to \texttt{loop\_types}; a line beginning with \texttt{repeat until} adds \texttt{repeat\_until}; and a line of the form \texttt{repeat (n)} adds \texttt{repeat} to \texttt{loop\_types} and the normalized count $n$ to \texttt{loop\_counts}.

Conditional structure is extracted from lines containing conditional contexts, such as \texttt{if <...>}, \texttt{wait until <...>}, and \texttt{repeat until <...>}. Within these contexts, SAC identifies coarse condition categories including \texttt{touching}, \texttt{edge}, \texttt{key\_press}, \texttt{mouse\_down}, \texttt{answer}, \texttt{equals}, \texttt{greater\_than}, and \texttt{less\_than}. SAC therefore captures condition types rather than full boolean-expression equivalence.

Named entities are extracted from Scratch-style bracketed arguments, such as \texttt{[score v]} in variable updates or \texttt{[message1]} in broadcast blocks. The extractor also handles common built-in quantities such as \texttt{x}, \texttt{y}, \texttt{size}, and \texttt{volume}. All names are normalized by lowercasing, trimming quotes, collapsing whitespace, and removing menu markers such as trailing \texttt{v}.

Numeric values are extracted after identifier masking. SAC first gathers recognized identifiers from event targets and name slots, masks them out of the text, and then extracts the remaining numeric material. This reduces spurious matches from identifiers that contain digits. Numeric normalization converts integer-valued floats such as \texttt{3.0} to \texttt{3} and otherwise keeps a compact decimal representation.

\paragraph{Natural-language-side extraction.}
Natural-language extraction is heuristic and relies on curated lexical inventories and regular-expression patterns. SAC lowercases the text and searches for cues corresponding to event triggers, actions, loops, conditions, numeric values, and named entities. Event extraction recognizes variants of green-flag triggers, key presses, sprite clicks, backdrop changes, and message reception. Action extraction matches verbal realizations of Scratch operations such as movement, appearance changes, variable updates, sound playback, waiting, asking, and broadcasting.

Loop and condition extraction similarly rely on lexical patterns. Phrases such as ``forever'', ``keep'', ``repeat'', and ``until'' are mapped to loop-related slots, while phrases expressing touching, key presses, mouse interaction, comparisons, answers, or edge contact are mapped to coarse condition types. Numeric values are recovered from explicit numeric mentions in the text.

\paragraph{Guided matching for names.}
Identifier-like slots are particularly prone to false negatives because natural language may refer to variables, costumes, backdrops, sounds, or broadcasts with minor surface variation. To reduce this problem, SAC uses pseudocode-guided matching for \texttt{variable\_names}, \texttt{costume\_names}, \texttt{backdrop\_names}, and \texttt{broadcast\_names}. Concretely, names extracted from the pseudocode are used as candidate identifiers on the natural-language side. This allows SAC to recover semantically important names even when they are embedded in more fluent descriptions.

\paragraph{Slot-wise scoring.}
After extraction, SAC compares corresponding slots with slot-specific scoring rules. Scalar slots such as \texttt{event\_type} and \texttt{event\_target} are scored by exact match. Set-valued slots such as \texttt{action\_types}, \texttt{loop\_types}, \texttt{condition\_types}, \texttt{numeric\_values}, and the various name slots are scored using set overlap as seen in Equation~\ref{eq:jaccard_score}.

\paragraph{Aggregate score.}
The final SAC score for a pair $(x,p)$ is the mean of the individual slot scores over all comparable slots as seen in Equation~\ref{eq:agg} In the main experiments, we use both the continuous SAC score and thresholded variants. We report perfect alignment when $\mathrm{SAC}=1.0$ and high alignment when $\mathrm{SAC}\geq0.85$.

\begin{table*}[t]
\centering
\small
\begin{tabular}{p{0.20\textwidth}p{0.32\textwidth}p{0.32\textwidth}c}
\toprule
\textbf{Slot} & \textbf{Extracted from NL} & \textbf{Extracted from pseudocode} & \textbf{Agree} \\
\midrule
\texttt{event\_type} 
& \{green\_flag\} 
& \{green\_flag\} 
& $\checkmark$ \\

\texttt{action\_types} 
& \{go\_to, move, set\_variable, turn\_left, turn\_right\} 
& \{go\_to, move, set\_variable, turn\_left, turn\_right\} 
& $\checkmark$ \\

\texttt{variable\_names} 
& \{score\} 
& \{score\} 
& $\checkmark$ \\

\texttt{loop\_types} 
& \{forever, repeat\_until\} 
& \{forever, repeat\_until\} 
& $\checkmark$ \\

\texttt{condition\_types} 
& \{edge, key\_pressed, touching\} 
& \{edge, key\_pressed, touching\} 
& $\checkmark$ \\

\texttt{numeric\_values} 
& \{-171, -97, 0, 10, 15\} 
& \{-171, -97, 0, 10, 15\} 
& $\checkmark$ \\
\bottomrule
\end{tabular}
\caption{Example SAC extraction for an aligned NL--pseudocode pair. The natural-language description is: ``When the green flag is clicked, set \texttt{score} to 0. Forever, go to $x=-171$, $y=-97$. If the space key is pressed, move 10 steps until you touch the edge or \texttt{Receiver1}. If the \texttt{a} key is pressed, turn right 15 degrees. If the \texttt{d} key is pressed, turn left 15 degrees.'' SAC assigns this pair a perfect alignment score of 1.00 because all comparable slots agree.}
\label{tab:sac_example}
\end{table*}

\paragraph{SAC variants.}
We use two SAC variants in evaluation. SAC$_{\text{nl}}$ compares the generated pseudocode $\hat{p}$ against the natural-language input $x$, measuring whether the generated program preserves the semantics expressed by the input. SAC$_{\text{ref}}$ compares the generated pseudocode $\hat{p}$ against the gold pseudocode $p$, measuring whether the generated program preserves the slot-level semantics of the reference program. These variants are complementary: SAC$_{\text{nl}}$ evaluates consistency with the user request, while SAC$_{\text{ref}}$ evaluates consistency with the benchmark reference.

\paragraph{Interpretation and limitations.}
SAC provides complementary information beyond surface-form metrics such as exact match or token F1 because it exposes which semantic components are preserved or violated. In particular, it can distinguish errors in event specification, loop structure, condition logic, numeric arguments, and entity references even when lexical overlap remains high. However, SAC is not a full semantic-equivalence test: it does not execute programs, and it may miss subtle ordering effects, interactions across scripts, or behavioral dependencies not represented in the slot schema. Its quality also depends on the coverage and robustness of the underlying rule-based extractors. We therefore treat SAC as a scalable semantic diagnostic rather than a complete substitute for execution-grounded evaluation.
\paragraph{Example extraction}
Table~\ref{tab:sac_example} shows an example of SAC extraction for a fully aligned NL--pseudocode pair. The example illustrates how SAC decomposes both sides into comparable behavioral slots before computing the aggregate alignment score.

\section{Training and Decoding Details}
\label{app:training_details}

\paragraph{Open-weight SFT.}

We fine-tune FLAN-T5, Qwen2.5-7B-Instruct, and Llama-3.1-8B-Instruct on the NL2Scratch training split using normalized scratchblocks pseudocode as the target representation. FLAN-T5 is fully fine-tuned in fp16 with AdamW, a learning rate of $10^{-4}$, and 2 training epochs. For Qwen2.5-7B-Instruct, we employ 4-bit QLoRA with bf16 computation using LoRA rank $r=64$, $\alpha=128$, and dropout $0.05$. Llama-3.1-8B-Instruct is fine-tuned using bf16 LoRA with the same adapter configuration. Both decoder-only models are trained for 2 epochs with a learning rate of $5\times10^{-5}$, cosine learning-rate decay, 3\% warmup, and an effective batch size of 16. LoRA adapters are applied to all attention and feed-forward projection layers ($q$, $k$, $v$, $o$, gate, up, and down projections). The maximum sequence length is 768 tokens for decoder-only models and 256 source / 256 target tokens for FLAN-T5. All experiments are conducted on NVIDIA GH200 and RTX 4090 GPUs.

\paragraph{Decoding and Candidate Selection.}
During inference, models generate up to 384 new tokens. The default baseline uses greedy decoding ($\texttt{num\_beams}=1$). For reranking experiments, we sample $N=8$ candidates using nucleus sampling ($T=0.7$, top-$p=0.9$). We evaluate five selection strategies: \textsc{first} (the first decoded candidate), \textsc{likelihood} (highest length-normalized log-likelihood), \textsc{SAC} (highest $\mathrm{SAC}_{nl}$ score), and two parser-constrained variants, \textsc{parse$\rightarrow$lik} and \textsc{parse$\rightarrow$SAC}, which first restrict candidates to parser-valid outputs before applying likelihood- or SAC-based ranking. If no candidate passes parser validation, selection falls back to the full candidate set.